\documentclass{article}

\usepackage{arxiv}

\usepackage[utf8]{inputenc} 
\usepackage[T1]{fontenc}    
\usepackage{hyperref}       
\usepackage{url}            
\usepackage{booktabs}       
\usepackage{amsfonts}       
\usepackage{nicefrac}       
\usepackage{microtype}      
\usepackage{lipsum}		
\usepackage{graphicx}
\usepackage[square,numbers]{natbib}
\usepackage{doi}
\usepackage{xcolor}

\usepackage{amsmath}
\usepackage{multirow}
\usepackage{float}
\usepackage{algorithm}
\usepackage{algpseudocode}
\usepackage{hhline}

\DeclareMathOperator*{\argmax}{arg\,max}
\DeclareMathOperator*{\argmin}{arg\,min}

\title{CovarNav: Machine Unlearning via Model Inversion and Covariance Navigation}

\author{$\text{Ali Abbasi}^*$
\And
$\text{Chayne Thrash}^*$
\And
Elaheh Akbari
\And
Daniel Zhang
\And
Soheil Kolouri
}

\date{\vspace{-8mm} Department of Computer Science, Vanderbilt University\\ \vspace{8mm} \tt {\{ali.abbasi, Chayne.thrash, elaheh.akbari, daniel.zhang.1, soheil.kolouri\}}\\@vanderbilt.edu \vspace{10mm}}	

\renewcommand{\undertitle}{}
\renewcommand{\shorttitle}{CovarNav: Machine Unlearning via Model Inversion and Covariance Navigation}
\let\svthefootnote\thefootnote
\newcommand\freefootnote[1]{%
  \let\thefootnote\relax%
  \footnotetext{#1}%
  \let\thefootnote\svthefootnote%
}

\hypersetup{
pdftitle={CovarNav: Machine Unlearning via Model Inversion and Covariance Navigation},
pdfauthor={Ali Abbasi, Chayne Thrash, Elaheh Akbari, Daniel Zhang, Soheil Kolouri},
pdfkeywords={Machine Unlearning, Continual Learning, Machine Learning},
}

\begin{document}
\maketitle

\begin{abstract}
	The rapid progress of AI, combined with its unprecedented public adoption and the propensity of large neural networks to memorize training data, has given rise to significant data privacy concerns. To address these concerns, machine unlearning has emerged as an essential technique to selectively remove the influence of specific training data points on trained models. In this paper, we approach the machine unlearning problem through the lens of continual learning. Given a trained model and a subset of training data designated to be forgotten (i.e., the ``forget set"), we introduce a three-step process, named CovarNav, to facilitate this forgetting. Firstly, we derive a proxy for the model's training data using a model inversion attack. Secondly, we mislabel the forget set by selecting the most probable class that deviates from the actual ground truth. Lastly, we deploy a gradient projection method to minimize the cross-entropy loss on the modified forget set (i.e., learn incorrect labels for this set) while preventing forgetting of the inverted samples. We rigorously evaluate CovarNav on the CIFAR-10 and Vggface2 datasets, comparing our results with recent benchmarks in the field and demonstrating the efficacy of our proposed approach.
\end{abstract}

\vspace{10mm}
\freefootnote{$^{*}$ Equal Contribution.}


\section{Introduction}

In light of the AI Revolution and the significant increase in public use of machine learning technologies, it has become crucial to ensure the privacy of personal data and offer the capability to erase or forget it from trained machine learning (ML) models on demand. This need is highlighted by many studies revealing risks, such as the ability to extract original data from models through model inversion attacks \cite{fredrikson2015model,zhang2020secret} or to identify if a particular sample was part of the training data through membership inference attacks \cite{shokri2017membership,hu2022membership}. Additionally, regulations like the European Union's General Data Protection Regulation (GDPR) \cite{GDRP}, California Consumer Privacy Act (CCPA) \cite{CCPA}, and PIPEDA
privacy legislation in Canada \cite{jaar2008canadian} stress the importance of individuals' control over their own data. 
More importantly, companies must now erase not just the data from users who have removed their accounts but also any models and algorithms developed using this data, e.g., \cite{federal2021california}. 

Erasing data from a model by retraining from scratch is computationally expensive, with significant economic and environmental implications. Consequently, Machine Unlearning has emerged as an active area of research. This field aims to efficiently remove specific data from trained systems without compromising their performance \cite{cao2015towards, golatkar2020eternal, bourtoule2021machine, graves2021amnesiac, gupta2021adaptive, sekhari2021remember, chen2023boundary, tarun2023fast}. In machine unlearning literature, terms like `removing,' `erasing,' and `forgetting' data refer to the process of completely obscuring a model’s understanding of sensitive data so that it cannot retain any meaningful information about it. Importantly, forgetting the target data set should minimally impact the model's performance on the remaining data. The critical question in machine unlearning, therefore, is how to forget a subset of data, such as a specific class, while retaining performance on the remaining data. This is particularly challenging when the entire training set is inaccessible, a practical assumption considering the growing size of training datasets and privacy considerations.

Similar to \cite{chen2023boundary}, in this paper, we focus on the problem of unlearning an entire class from
deep neural networks (DNNs).  Our proposed method is inspired by the close relationship between the fields of continual learning \cite{de2019continual, kudithipudi2022biological} and machine unlearning. Specifically, continual learning approaches aim to prevent `forgetting' in machine learning models, preserving their performance on previous tasks while learning new ones. In the context of machine unlearning, methods used in continual learning to minimize forgetting can be adapted to: 1) maximize forgetting on the target set, and 2) minimize forgetting on the remainder of the data. This interrelation has also been noted and influenced recent works in machine unlearning \cite{carta2022ex, yan2022generative, madaan2023heterogeneous}. Recently, a class of gradient-projection-based continual learning algorithms, which ensure performance preservation on previous tasks (i.e., zero backward transfer), has been proposed \cite{saha2020gradient, wang2021covarnull, abbasi2022sparsity}. Although these methods are often critiqued in the continual learning context for not allowing positive backward transfer, this aspect renders them ideal for machine unlearning problems, where zero backward transfer is desirable. Hence, in this paper, we adopt a similar approach to that of Saha et al. \cite{saha2020gradient} and Wang et al. \cite{wang2021covarnull} to preserve performance on the remaining data while effectively forgetting the target set. Notably, however, we do not assume access to the training set.

In our unlearning setting, we assume access to the target set, i.e., the set to be forgotten, and the model, but not to the rest of the training set. However, it is essential to note that a substantial body of research on model inversion attacks \cite{yin2020dreaming, wang2021variational} exploits the deep DNNs' tendency to memorize to approximate the model's training set. In this paper, we propose to utilize model inversion to approximate the training data of the model's remaining classes. Our goal is to preserve the performance of the model on these inverted data while effectively forgetting the target set. In our ablation studies, we compare this strategy against scenarios where we have access to the training set, demonstrating the effectiveness of model inversion.

\noindent \textbf{Our specific contributions} in this paper are as follows:
\begin{itemize}
    \item Introduced a novel machine unlearning framework named Covariance Navigation (CovarNav) for forgetting an entire class of data from the training and preserving performance on the remaining data while not having access to the training data.     
    \item Demonstrate that CovarNav provides superior results compared to state-of-the-art machine unlearning approaches on various benchmark datasets, excelling in forgetting the target set and preserving performance on the remaining set.
    \item Performed extensive ablation studies to demonstrate the contribution of all proposed steps to the final accuracy. 
\end{itemize}

\begin{figure}[t!]
    \centering
    \includegraphics[width=\columnwidth]{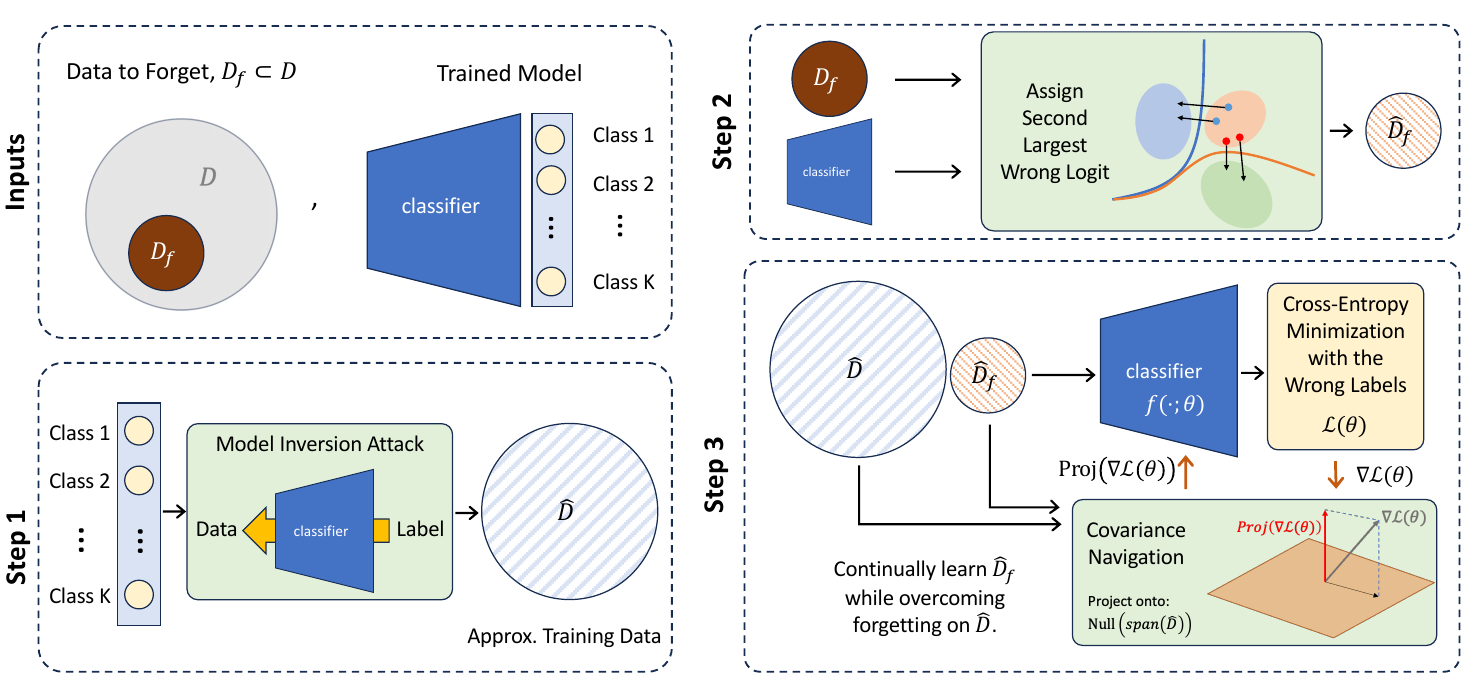}
    \caption{The overview of our proposed machine unlearning framework is presented. Given a dataset, $D_f$, to be forgotten, and a trained model, $f(\cdot;\theta)$, we first perform a model inversion attack to obtain a proxy for the training data, denoted as $\hat{D}$. Following the prior work in the literature, we replace the labels of $D_f$ with those of the closest neighboring class (specifically, the second largest logit) to produce $\hat{D}_f$. The network is then trained on $\hat{D}_f$ using gradient projection into the null space of $\hat{D}$'s neural activations, ensuring that the rest of the data remains unaffected.}
    \label{fig:overview}
\end{figure}

\section{Related Work}

\noindent \textbf{Machine Unlearning} aims to erase specific training data from a pre-trained model, a process also known as `forgetting,' while ensuring minimal impact on the model's overall performance on the remaining data \cite{xu2023machine,zhang2023review,nguyen2022survey}.  The term machine unlearning was coined by Cao \& Yang \cite{cao2015towards}, while the core concept can be tracked in the literature before this work \cite{cauwenberghs2000incremental,romero2007incremental,tsai2014incremental}. Machine unlearning methods can generally be categorized into exact unlearning and approximate unlearning approaches. Exact unlearning methods ensure that the data distributions in both a natively retrained model, i.e., a model trained from scratch on the remaining data, and a model that has undergone unlearning are indistinguishable \cite{cao2015towards,cao2018efficient,brophy2021machine,bourtoule2021machine,liu2021federaser,jose2021unified,kashef2021boosted,ullah2021machine}. Unfortunately, exact unlearning is only feasible for simpler, well-structured models. Thus, approximate unlearning approaches have been developed for more complex models, including diverse types of deep neural networks \cite{golatkar2020eternal,neel2021descent,graves2021amnesiac,thudi2022unrolling,chen2023boundary}. Moreover, machine unlearning could be achieved through data reorganization \cite{cao2015towards,bourtoule2021machine,tarun2023fast}, e.g., pruning and obfuscation, and model manipulation \cite{golatkar2020eternal,sekhari2021remember,graves2021amnesiac,thudi2022unrolling,chen2023boundary}. 

In this paper, we devise an approximate unlearning approach based on model manipulation, which involves altering the trained model's parameters and only requires access to the trained model and the `forget data.' Our work is closely related to Boundary Shrink \cite{chen2023boundary} but differs significantly in technical aspects. In particular, and in contrast to \cite{chen2023boundary}, we utilize `model inversion' to construct a proxy for the training data and apply principles from the continual learning research community to effectively forget the targeted data, i.e., the forget data, while preserving the model's performance on the remaining data.

\noindent \textbf{Continual learning} deals with learning from a stream of data or tasks while 1) enhancing backward knowledge transfer, which aims to maintain or improve performance on previously learned tasks, thereby mitigating catastrophic forgetting, and 2) bolstering forward knowledge transfer, where learning a current task can boost performance on or reduce the learning time for future tasks \cite{de2021continual,kudithipudi2022biological}. Addressing `catastrophic' forgetting is at the heart of continual learning approaches. Current strategies to address this problem broadly fall into three categories: 1) Memory-based methods, which include techniques like memory rehearsal/replay, generative replay, and gradient projection \citep{shin2017continual,farquhar2018towards,rolnick2019experience,rostami2020generative,farajtabar2020orthogonal}; 2) Regularization-based approaches that impose penalties on parameter alterations crucial to previous tasks \citep{kirkpatrick2017overcoming,zenke2017continual,aljundi2017memory,kolouri2020sliced,li2021lifelong,von2019continual}; and 3) Architectural methods focusing on model expansion, parameter isolation, and masking \citep{schwarz2018progress,mallya2018packnet,mallya2018piggyback,wortsman2020supermasks,ben-iwhiwhu2023lifelong}. Recently, methods based on gradient projection \citep{farajtabar2020orthogonal,saha2020gradient,deng2021flattening,wang2021training,lin2022trgp,abbasi2022sparsity,zhao2023CVPR} have demonstrated remarkable performance while providing an elegant theoretical foundation for overcoming forgetting in continual learning.

Interestingly, continual learning is closely related to machine unlearning \cite{shibata2021learning,heng2023selective,zhang2023machine}. Whether the goal is to prevent performance degradation on retained data while forgetting target data or to pinpoint critical parameters essential for effective unlearning, techniques developed in the domain of continual learning are increasingly being recognized as valuable tools for machine unlearning. These techniques offer insights into how models can be adapted dynamically, balancing retaining old information with acquiring or removing new knowledge. In this paper, we propose a gradient projection framework that is similar to \cite{saha2020gradient,wang2021covarnull,abbasi2022sparsity,lin2022trgp,zhao2023CVPR} to unlearn the target data while maintaining performance on the retained data. We denote this gradient projection algorithm as Covariance Navigation, leading to our proposed method, CovarNav.

\noindent \textbf{Model Inversion} \cite{fredrikson2015model,veale2018algorithms} refers to attack strategies that aim to reconstruct training data or infer sensitive attributes or details from a trained model. These methods typically involve optimizing inputs in the data space to maximally activate specific output neurons (e.g., target classes). However, this optimization is inherently ill-posed due to the many-to-one mapping characteristic of deep neural networks—where a variety of inputs can lead to the same output. The existing literature proposes multiple types of priors (i.e., regularizations) to make the problem more tractable. Such regularizations range from simpler techniques like Total Variation and image norm \cite{mahendran2015understanding,mordvintsev2015deepdream} to more complex methods involving feature statistics \cite{yin2020dreaming} and generative models \cite{wang2021variational}.

Model inversion has found significant applications in continual learning \cite{carta2022ex,yan2022generative,madaan2023heterogeneous} and machine unlearning \cite{chundawat2023zero}, serving as a tool for data reconstruction and model privacy evaluation. In this paper, we employ a model inversion attack to construct a representative dataset—acting as a proxy for the retained data—which enables us to preserve the network's performance on this data while selectively removing or `forgetting' the target set.

\section{Method}

This section outlines the machine unlearning setting in which we operate and introduces our proposed framework, CovarNav. We begin by establishing our notations and then detail our three proposed steps, as illustrated in Figure \ref{fig:overview}. 

Let $D = \{(x_i, y_i)\}_{i=1}^N\subseteq \mathcal{X}\times \mathcal{Y}$ represent the private training dataset where $x_i\in \mathcal{X}$ denotes the input (e.g., images) and $y_i\in \mathcal{Y}=\{1,\cdots,K\}$ denotes its corresponding label. Here, $\mathcal{X}$ and $\mathcal{Y}$ represent the input and label spaces, respectively. Let $f(\cdot;\theta):\mathcal{X}\rightarrow \mathcal{Y}$ denote the DNN classifier, with parameters $\theta$, and let $\theta^*$ denote the optimal parameters of the model after being trained on dataset $D$. We denote the target set, i.e., the forgetting data as $D_f\subset D$, and use $D_r=D\backslash D_f$ to represent the remainder of the data, where performance retention is required.
The primary objective in our machine unlearning setting is to update  \(\theta^*\) to: 1) degrade the model's performance on \(D_f\), and 2) maintain performance on $D_r$ while achieving this without utilizing the data in $D_r$ during the unlearning process.


We propose CovarNav as a robust solution to the unlearning problem that operates post hoc, requiring no stored statistics from $D_r$, and eliminates the need for users to anticipate future requests for data forgetting. CovarNav consists of three core steps: 1) employing model inversion attack to procure pseudo samples of \(D_r\), which we denote as $\hat{D}_r$, 2) constructing a forgetting objective based on $D_f$, and 3) optimizing the forgetting objective with gradient-projection to preserve the model's performance on \(\hat{D}_r\). Next, we delve into these three steps and formalize them.


\subsection{Model Inversion}
Despite the lack of access to \(D_r\), the trained model \(f(\cdot;\theta^*)\) retains important information about the original dataset. We propose utilizing model inversion attacks, which exploit this retained information, to derive pseudo samples for \(D_r\) and construct a proxy set $\hat{D}_r$, representing the remaining data.

Let \(c_f\) denote the class id we intend to forget and define the set of remaining labels as \(\mathcal{Y}_r = \mathcal{Y}\backslash\{c_f\}\). In line with the work of  Yin et al. \cite{yin2020dreaming}, we formulate the model inversion attack for a batch of target labels $\{y_j\in \mathcal{Y}_r\}_{j=1}^B$ as:

\begin{align}\label{inv_obj}
    \hat{D}_r = \argmin_{\{x_j\in \mathcal{X}\}_{j=1}^B} ~~ \sum_{j=1}^B \big( \mathcal{L}_{\text{task}}(x_j, y_j,\theta^*) + \mathcal{R}_{\text{prior}}(x_j) \big)+ 
    \alpha_f \mathcal{R}_{\text{feat}}(\{x_i\}_{i=1}^B, \theta^*),
\end{align}
where $\mathcal{L}_{\text{task}}$ is the classification loss (e.g., cross-entropy), \(\mathcal{R}_{\text{prior}}\) is an image regularization term that acts as a weak prior for natural images \cite{mordvintsev2015deepdream}, and \(\mathcal{R}_{feat}\) is a feature-statistics loss as used in \cite{yin2020dreaming}. In particular, for \(\mathcal{R}_{\text{prior}}(x)\) we use the following regularization terms:
\begin{align}
    \mathcal{R}_{\text{prior}}(x) = \alpha_{\text{TV}}\mathcal{R}_{\text{TV}}(x)+ \alpha_{\ell_2}\mathcal{R}_{\ell_2}(x),
\end{align}
where $\mathcal{R}_{TV}(x)$ denotes the total variation of image $x$, $\mathcal{R}_{\ell_2}(x)$ is the $\ell_2$ norm of the image, and $\alpha_{\text{TV}},\alpha_{\ell_2},\alpha_f>0$ are the regularization coefficients. The feature-statistics regularization \(\mathcal{R}_{feat}\) utilizes the fact that many modern DNNs incorporate batch normalization \cite{ioffe2015batch} to accelerate and stabilize training, and the fact that the batch normalization layers contain the running mean and variance of training data.  Hence,  \(\mathcal{R}_{feat}\) cleverly employs this running mean and variance and requires the inverted sample $\{x_{j}\}_{j=1}^B$ to follow the same feature statistics via:
\begin{align}
\mathcal{R}_{\text{feat}}(\{x_i\}_{i=1}^B) = \sum_l \left\| \mu_l(\{x_i\}_{i=1}^B) - m_l) \right\|_2 + \sum_l \left\| \sigma_l^2(\{x_i\}_{i=1}^B) - v_l \right\|_2.
\end{align}
Here, \(m_l\) and \(v_l\) are the saved means and variances at the \(l^{\text{{th}}}\) batch normalization layer, and $\mu_l$ and $\sigma^2_l$ are the corresponding mean and variance for the set $\{x_i\}_{i=1}^B$.

\subsection{Forgetting Objective}\label{sec:forgetobj}

We aim to update the parameters $\theta^*$ such that the model, $f(\cdot;\theta)$, forgets the set, $D_f$. However, there are multiple ways of formalizing this forgetting process. For instance, one approach is to define the forgetting objective as maximizing the cross-entropy loss on $D_f$. Alternatively, one could assign random incorrect labels to the samples in $D_f$ and minimize the cross-entropy loss for these incorrect labels. Instead of assigning wrong labels randomly, Chen et al. \cite{chen2023boundary} proposed to find the closest wrong class through untargeted evasion attacks on \(D_f\)'s samples via Fast Gradient Sign Method (FGSM) \cite{goodfellow2014explaining}. In this paper, we follow a similar rationale to that of \cite{chen2023boundary}; however, we show that instead of using an untargeted evasion attack to mislabel \(D_f\), one can mislabel samples based on their largest wrong logit according to $f(\cdot;\theta^*)$. Through ablation studies, we show that this strategy is at least as effective as the one used in \cite{chen2023boundary}. We denote this mislabeled forget set as $\hat{D}_f=\{(x_{f,j},\hat{y}_{j,f})\}_{j=1}^{N_f}$ where $\hat{y}_{j,f}$ correspond to the wrong class with the largest logit. Finally, we set up the forgetting problem as: 
\begin{align}
    \argmin_\theta \sum_{j=1}^{N_f} \mathcal{L}_{\text{task}}(x_{f,j},\hat{y}_{f,j},\theta).
    \label{eq:forget}
\end{align}
Note that, minimizing $\theta$ according to the above optimization problem leads to forgetting $D_f$, however, at the expense of losing performance on $D_r$, i.e., catastrophic forgetting on $D_r$. Next, we discuss our strategy for avoiding catastrophic forgetting on $D_r$ while solving \eqref{eq:forget}.



\subsection{Covariance Navigation}
Let \(\theta^*\) denote the network's original parameters, and \(\theta\) represent the parameters after forgetting \(D_f\). Then, ideally, we would like \(f(x_r;\theta) = f(x_r;\theta^*)\) for \(\forall x_r \in D_r\). To achieve this, we follow the work of Saha et al. \cite{saha2020gradient} and Wang et al. \cite{wang2021covarnull}, which we briefly describe here. With abuse of notation, we denote the network's activations for input \(x_r\) at layer \(l\) as \(x_r^l\). Moreover, let us denote the original network's weights at layer \(l\) as \(W^l\) and the updated weights as \(W^l_f = W^l + \Delta W_f^l\). It should be clear that one way to enforce \(f(x_r;\theta) = f(x_r;\theta^*)\) is to require the network activations at each layer be preserved, i.e., 
\begin{align}
    W^l x^l_r = W_f^l x^l_r,
\end{align}
for $\forall x_r \in D_r$. Hence, we can immediately see that for the above equation to be valid, we require 
\begin{align}
    \Delta W_f^l x_r^l=0,~~~\forall x_r\in D_r.
    \label{eq:nullspace}
\end{align}
Eq \ref{eq:nullspace} implies that if the gradient updates at each layer are orthogonal to the activations of all the data in $D_r$, i.e., $X^l_r = \{x^l_{r, i}\}_{i=1}^{N_r}$, then the network is guaranteed to satisfy \(f(x_r;\theta) = f(x_r;\theta^*)\). Hence, by projecting the gradient updates onto $\text{Null}(X^l_r)$ for each layer $\forall l$, we can guarantee no performance loss on $D_r$. Moreover, it is straightforward to confirm that the null space of $X_r^l$ is equal to the null space of the uncentered feature covariance, i.e., $S^l_r=X^l_r(X_r^{l})^T$:
\begin{align}
    \text{Null}(X^l_r) = \text{Null}(S_r^l)    
\end{align}



Thus, we can alternatively project the gradient updates onto the null space of the covariance of the activations; this method is what we refer to as `Covariance Navigation.'

We note that $\text{Null}(S_r^l)$ could be empty, for instance, when $S_r^l$ has many small eigenvalues that are all non-zero. An empty null space indicates that the gradient updates are mapped to zero, and in other words, we would not be able to forget $D_f$. To avoid such a scenario, an approximate null space of $S_r^l$ is utilized. Let $\{\lambda_i\}_{i=1}^{d_l}$ denote the eigenvalues of $S_r^l$ sorted in a descending manner. Then, to obtain the approximate null space, we calculate: 
\begin{align}
\rho_k= \frac{\sum_{i=1}^k \lambda_i}{\sum_{j=1}^{d_l} \lambda_j}    
\end{align}
and find the smallest $k$ that satisfies $\rho_k\geq p$ where $p\in[0,1]$ is a hyperparameter for approximating the null space, and set $\lambda_{k+1:}$ to zero, leading to a $(d_l-k)$-dimensional null space. Note that by reducing $p$, the null space expands but at the cost of a possible increase in forgetting for $D_r$. Lastly, since we do not have access to $D_r$ during the unlearning phase, use the inverted set $\hat{D}_r$ as a proxy for this dataset.





\begin{algorithm}[h!]
\caption{CovarNav Unlearning Algorithm}
\hspace*{\algorithmicindent} \textbf{Inputs} Forget data $D_f$, Trained Model $h(\cdot;\theta^{*})$, lr $\tau$
\begin{algorithmic}[1]
\Procedure{CovarNav}{}
    \State \textcolor{blue}{// Step 1: Perform Model Inversion}
    \State Uniformly sample $\{y_{j} \in \mathcal{Y}_r\}_{j=1}^{B}$ 
    \State Obtain $\hat{D}_{r}$ from \ref{inv_obj} \\
    \State \textcolor{blue}{// Step 2: Mislabel $D_f$}    
    \State $\hat{D}_{f} \leftarrow \emptyset$
    \For{$x_f, y_f$ from $D_f$}
        \State $\hat{y}_f \leftarrow \argmax_{\{i\in \mathcal{Y}_r\}} [h(x;\theta^{*})]_i$
        \State $\hat{D}_{f} \leftarrow \hat{D}_{f} \cup \{(x_f, \hat{y}_f)\}$
    \EndFor \\
    \State \textcolor{blue}{// Step 3: Covariance Navigation} 
    \State $S^l_r \leftarrow X^l_r(X_r^{l})^T$
    \State Compute $Null(S_{r}^{l})$
    \State $\theta \leftarrow \theta^*$
    \For{$e$ in Epochs}
        \State $\text{L} \leftarrow \sum_{j=1}^{N_f} \mathcal{L}_{\text{task}}(x_{f,j},\hat{y}_{f,j},\theta)$ 
        \State $g \leftarrow \nabla_\theta\text{L}$
        \State $\theta \leftarrow \theta -   Proj_{\text{Null}}[\text{Adam}(g, \tau)]$
    \EndFor
    \\
    \State Return $\theta$
\EndProcedure
\end{algorithmic}
\end{algorithm}



\section{Experiments}

In this section, we present experiments conducted on two prominent machine unlearning benchmark datasets: CIFAR-10 and VGGFace2. These experiments aim to assess the efficacy of our proposed model. We implemented all experiments and baselines using Python 3.8 and the PyTorch library, on an NVIDIA RTX A5000 GPU. 

\subsection{Datasets}
Following the recent work in the literature \cite{chen2023boundary,tarun2023fast,lee2023undo}, we conduct experiments on CIFAR-10 \cite{krizhevsky2009learning} and VGGFace2 \cite{DBLP:journals/corr/abs-1710-08092} datasets. CIFAR-10 contains 10 classes of 32 x 32 images, with a total of 50,000 and 10,000 images for training and testing sets, respectively. For VGGFace2, we follow the procedure outlined in \cite{golatkar2020eternal} to create a set containing 10 faces with 4587 training and 1000 test samples. 

\subsection{Metrics}

To evaluate the efficacy of a method for unlearning, it's crucial that the model, post-unlearning, holds minimal information about the data intended to be forgotten while still maintaining its performance on the data that is retained. In this context, our primary metrics involve measuring the model's accuracy both before and after the unlearning process. This measurement is conducted on the training and testing subsets of both the data to be forgotten and the data to be retained, denoted as $D_f$ and $D_r$ for the training sets, and $D_{ft}$ and $D_{rt}$ for the testing sets, respectively. Consequently, the ideal outcome would be a reduced accuracy on $D_f$ and $D_{ft}$ (optimally reaching zero), alongside maintaining or improving accuracy on $D_r$ and $D_{rt}$.

While accuracy measurements on $D_f$, $D_{ft}$, $D_r$, and $D_{rt}$ are informative, they can be misleading when used in isolation. This is largely because achieving low accuracy on $D_f$ and $D_{ft}$ could simply be the result of adjusting the classifier's weights. Addressing this concern, several studies \cite{golatkar2020eternal, tarun2023fast, chundawat2023zero} have proposed different metrics that incorporate relearn time to more accurately assess the effectiveness of machine unlearning methods. The idea behind these approaches is that the speed of relearning the forgotten dataset ($D_f$) reflects the residual information in the model, thereby evaluating the unlearning algorithm's thoroughness. Notably, the Anamnesis Index proposed by Chundawat et al. \cite{chundawat2023zero} stands out as it calculates the time required for a model $M$ to achieve $\alpha\%$ of the original model $M_{\text{orig}}$'s accuracy on $D_f$.  In short, let the number of mini-batches (steps) required
by a model $M$ to come within $\alpha\%$ range of the accuracy of $M_{\text{orig}}$ on the forget dataset ($D_f$) be denoted as $r_t(M, M_{\text{orig}}, \alpha)$. For $M_u$ and $M_s$ denoting the unlearned model and the model trained from scratch on $D_r$, the Anamnesis Index (AIN) \cite{chundawat2023zero} is defined as: 
\begin{align}
    \text{AIN}(\alpha)= \frac{r_t(M_u, M_{\text{orig}}, \alpha)}{r_t(M_s, M_{\text{orig}}, \alpha)}.
\end{align}
 AIN ranges from $0$ to $+\infty$, with $AIN=1$ indicating an ideal unlearning algorithm. AIN values significantly lower than $1$ imply that the model retains information about the classes it was supposed to forget. This lower value also suggests that the model quickly reacquires the ability to make accurate predictions on the forgotten classes. Such a scenario often occurs when modifications to the model, particularly in its final layers, temporarily impair its performance on the forgotten classes, but these changes are easily reversible. On the other hand, an AIN value considerably higher than $1$ might indicate that the unlearning process involved substantial alterations to the model's parameters. These extensive changes are so pronounced that they make the unlearning process apparent. Following the suggested values for $\alpha$ in \cite{chundawat2023zero}, in this paper, we use $\text{AIN}(\alpha=0.1)$ as our complementary metric to the accuracy.



\setlength{\tabcolsep}{5pt}
\begin{table*}[t]
    \centering
    \hspace*{-10pt}\begin{tabular}{c|cccccccc}
        \hline
         & \shortstack{Method\\~} & \shortstack{Post-train\\~} & \shortstack{\\No Access\\ to \(D_r\)} & \shortstack{$Acc_{D_{r}}\uparrow$\\~} & \shortstack{$Acc_{D_{f}}\downarrow$\\~} & \shortstack{$Acc_{D_{rt}}\uparrow$\\~} & \shortstack{$Acc_{D_{ft}}\downarrow$\\~} & \shortstack{AIN \\ ($\alpha=0.1$)}\\
         \hline
         \multirow{12}{*}{\rotatebox[origin=c]{90}{CIFAR-10}} & Original & N/A & N/A & $99.43$ & $99.78$ & $88.37$ & $91.1$ & N/A\\
         & Retrain on \(D_r\) & $\checkmark$ & $\times$ & $99.5$ & $0.0$ & $86.46$ & $0.0$ & $1.0$\\
         & Finetune \cite{golatkar2020eternal} & $\checkmark$ & $\times$ & $99.13$ & $0.0$ & $80.86$ & $0.0$ & $6.39$\\ 
         & Negative Gradient \cite{golatkar2020eternal} & $\checkmark$ & $\times$ & $94.65$ & $0.0$ & $83.64$ & $0.0$ & $7.73$\\
         & Amnesiac* \cite{graves2021amnesiac} & $\times$ & $\checkmark$ & $60.49$ & $0.0$ & $51.5$ & $0.0$ & $42.6$\\
         & ERM-KTP* \cite{Lin_2023_CVPR} & $\times$ & $\times$ & $98.36$ & $0.0$ & $87.95$ & $0.0$ & $8.76$\\
         & \(D_f\) w/ Random Labels \cite{golatkar2020eternal} & $\checkmark$ & $\checkmark$ & $96.04$ & $0.83$ & $85.79$ & $0.63$ & $17.03$\\
         & Boundary Shrink \cite{chen2023boundary} & $\checkmark$ & $\checkmark$ & $96.91$ & $0.29$ & $86.92$ & $0.4$ & $6.41$\\ 
         & Maximize \(D_f\) Entropy & $\checkmark$ & $\checkmark$ & $92.82$ & $19.87$ & $81.68$ & $16.07$ & $45.59$\\
         & Largest Wrong Logit & $\checkmark$ & $\checkmark$ & $99.2$ & $0.0$ & $88.98$ & $0.03$ & $6.44$\\
         & Largest Wrong Logit + $||\Delta \theta||^2$ & $\checkmark$ & $\checkmark$ & $99.22$ & $0.0$ & $88.98$ & $0.07$ & $11.49$\\
         \hline
         & \shortstack{\\CovarNav (Ours)} & $\checkmark$ & $\checkmark$ & $99.27$ & $0.0$ & $89.02$ & $0.03$ & $8.52$\\
         \hhline{=========}
         \multirow{12}{*}{\rotatebox[origin=c]{90}{VGGFace2}} & Original & N/A & N/A & $100.0$ & $100.0$ & $80.89$ & $90.0$ & N/A\\
         & Retrain on \(D_r\) & $\checkmark$ & $\times$ & $100.0$ & $0.0$ & $76.7$ & $0.0$ & $1.0$\\
         & Finetune \cite{golatkar2020eternal} & $\checkmark$ & $\times$ & $99.68$ & $0.0$ & $75.81$ & $0.0$ & $28.41$\\ 
         & Negative Gradient \cite{golatkar2020eternal} & $\checkmark$ & $\times$ & $94.75$ & $0.0$ & $70.11$ & $0.0$ & $15.94$\\ 
         & Amnesiac* \cite{graves2021amnesiac} & $\times$ & $\checkmark$ & $12.88$ & $0.0$ & $11.11$ & $0.0$ & $0.01$\\
         & ERM-KTP* \cite{Lin_2023_CVPR} & $\times$ & $\times$ & $100.0$ & $0.0$ & $78.96$ & $0.0$ & $7.95$\\ 
         & \(D_f\) w/ Random Labels \cite{golatkar2020eternal} & $\checkmark$ & $\checkmark$ & $99.65$ & $0.0$ & $77.89$ & $1.0$ & $151.67$\\ 
         & Boundary Shrink \cite{chen2023boundary} & $\checkmark$ & $\checkmark$ & $99.43$ & $0.0$ & $78.15$ & $0.0$ & $74.67$\\ 
         & Maximize \(D_f\) Entropy & $\checkmark$ & $\checkmark$ & $99.62$ & $8.89$ & $78.04$ & $8.67$ & $151.67$\\ 
         & Largest Wrong Logit & $\checkmark$ & $\checkmark$ & $99.94$ & $0.0$ & $77.93$ & $0.0$ & $25.59$\\ 
         & Largest Wrong Logit + $||\Delta \theta||^2$ & $\checkmark$ & $\checkmark$ & $99.95$ & $0.0$ & $78.15$ & $0.0$ & $25.59$\\ 
         \hline
         & \shortstack{\\CovarNav (Ours)} & $\checkmark$ & $\checkmark$ & $100.0$ & $0.0$ & $80.96$ & $3.0$ & $32.02$\\ 
         \hline
    \end{tabular}
    \caption{Performance comparison between baselines and CovarNav on both CIFAR-10 and VGGFace2. Asterisks denote that the unlearning method was applied to a different original model due to having to change the training procedure.}
    \label{tab:vgg_results}
\end{table*}

\subsection{Baselines}
To assess the quality of our proposed framework, we compared our method with the following baselines.

\textbf{Retrain.} This baseline involves training the model from scratch solely using the retained dataset \(D_r=D \backslash D_f\). While time-consuming and inefficient, this approach serves as a benchmark to evaluate the effectiveness of any unlearning model, and is essential for understanding the impact of unlearning on the model's performance.

\textbf{Finetune.} \cite{golatkar2020eternal} For this baseline, we finetune the original trained model on \(D_r\) with a large learning rate, which acts as a scrubbing procedure for $D_f$. 

\textbf{Negative Gradient.} \cite{golatkar2020eternal} We finetune the original model on the entire dataset. However, we maximize the loss for data samples corresponding to \(D_f\) and minimize the loss for samples corresponding to \(D_r\). We clamp the loss to chance level to prevent divergence. This aims to damage features predicting $D_f$ correctly while maintaining high performance on $D_r$.

\textbf{Amnesiac Unlearning.} \cite{graves2021amnesiac} The amnesiac unlearning method is a training-time algorithm, meaning it operates during, not after, the training phase. It records parameter updates across multiple batches during the initial training. This approach is effective in maintaining accuracy, especially when the number of batches with samples from \(D_f\) is limited. Consequently, we sub-sample the original dataset to include only a select number of batches containing forgetting data. Throughout the training, we save the gradient updates corresponding to these forgetting batches and later reverse these updates. 

\textbf{ERM-KTP.} \cite{Lin_2023_CVPR} ERM-KTP is another training-time algorithm that adds a mask layer to the original model, which learns the relationships between features and classes and also enforces that features have limited usage in multiple classes. This requires the model to be trained initially with this masking layer. After training, the features related to $D_f$ are removed, and the model is fine-tuned on \(D_r\) without labels to ensure consistency with the original model.

\textbf{\(D_f\) with Random Labels} \cite{golatkar2020eternal} This baseline changes the forgetting objective from maximizing the cross-entropy loss on $D_f$ to first assigning random wrong labels to samples from $D_f$ and then minimizing the cross-entropy loss to these wrong random labels. 

\textbf{Boundary Shrink} \cite{chen2023boundary} Similar to the Random Labels baseline, this recent approach also implements a forgetting objective. It starts by assigning incorrect labels to samples from \(D_f\) and then focuses on minimizing the cross-entropy loss for these mislabeled samples. In the Boundary Shrink method, the mislabeling is achieved by applying FGSM \cite{goodfellow2014explaining} to samples from \(D_f\), aiming to locate the nearest decision boundary and the nearest wrong class.

In addition to the existing methods in the literature, and to better understand the effect of forgetting objectives, we introduced three additional baselines described below. 

\textbf{Maximize \(D_f\) Entropy} This baseline maximizes the cross-entropy loss on \(D_f\), increasing the entropy of the output distribution for the forgetting data.

\textbf{Largest Wrong Logit} This baseline implements a forgetting objective that assigns the largest incorrect logit as the label to samples from $D_f$. It then minimizes the cross-entropy loss on the mislabeled $D_f$ samples.

\textbf{Largest Wrong Logit + $||\Delta \theta||^2$} The same as the previous baseline but with an additional regularization term on the change in weights between the unlearned and the original model. The additional $\ell_2$ regularization is expected to reduce forgetting on $D_r$.

\subsection{Experiment Settings}
In our study, we employ the ResNet-18 model, as outlined in He et al., \cite{He2015}, for all experiments.
For the CIFAR-10 experiments, we utilize the pre-trained weights made available for the Google 2023 Machine Unlearning challenge\footnote{\url{https://storage.googleapis.com/unlearning-challenge/weights_resnet18_cifar10.pth}}. With respect to VGGFace2, our methodology aligns with that of \cite{golatkar2020eternal}; initially, we pre-train our model on a dataset comprising 100 faces, followed by fine-tuning on a smaller dataset containing only 10 faces. During both the pre-training and fine-tuning phases, we use Stochastic Gradient Descent (SGD) and train for 100 epochs. The training settings include a learning rate of 0.01, a momentum of 0.9, and a weight decay factor of 1e-4.

When training CovarNav, we use the Adam \cite{KingBa15} optimizer. For CIFAR-10, we unlearn for 25 epochs with a learning rate of $1e-5$ and set $p = 1.0$. For VGGFace2, we train for 100 epochs with a learning rate of $1e-4$ and set $p = 0.9$. For both datasets, we create \(\hat{D}_{r}\) with 100 samples per class.

\begin{table*}[!t]
    \centering
    \begin{tabular}{c|cccccc}
        \hline
        & Dataset for Covariance & Dataset Size & $Acc_{D_{r}}\uparrow$ & $Acc_{D_{f}}\downarrow$ & $Acc_{D_{rt}}\uparrow$ & $Acc_{D_{ft}}\downarrow$\\
        \hline
         \multirow{2}{*}{CIFAR-10} & \(D_r\) & $45000$ & $99.22$ & $0.0$ & $88.78$ & $0.0$\\
         & \(\hat{D}_{r}\) & $900$ & $99.27$ & $0.0$ & $89.02$ & $0.03$\\
         \hline
         \multirow{2}{*}{VGGFace2} & \(D_r\) & $4167$ & $100.0$ & $0.0$ & $80.93$ & $3.0$\\
         & \(\hat{D}_{r}\) & $900$ & $100.0$ & $0.0$ & $80.96$ & $3.0$\\
         \hline
    \end{tabular}
    \caption{Effect of using inverted retained dataset, $\hat{D}_{r}$, for covariance matrix instead of the actual data $D_r$.}
    \label{tab:ablate_inversion}
\end{table*}

\begin{table*}[!t]
    \centering
    \begin{tabular}{c|ccccc}
        \hline
        & Method & $Acc_{D_{r}}\uparrow$ & $Acc_{D_{f}}\downarrow$ & $Acc_{D_{rt}}\uparrow$ & $Acc_{D_{ft}}\downarrow$\\
        \hline
         \multirow{2}{*}{CIFAR-10} & Maximize Entropy & $97.0$ & $1.89$ & $85.71$ & $1.27$\\
         & Random Labels & $96.12$ & $0.01$ & $85.29$ & $0.03$\\
         & Boundary Shrink \cite{chen2023boundary} & $97.75$ & $0.03$ & $87.49$ & $0.1$\\
         & Largest Wrong Logit (Ours) & $99.27$ & $0.0$ & $89.02$ & $0.03$\\
         \hline
         \multirow{2}{*}{VGGFace2} & Maximize Entropy & $99.98$ & $6.83$ & $80.44$ & $10.0$\\
         & Random Labels & $99.91$ & $0.63$ & $78.67$ & $3.33$\\
         & Boundary Shrink \cite{chen2023boundary} & $100.0$ & $0.0$ & $80.11$ & $1.0$\\
         & Largest Wrong Logit (Ours) & $100.0$ & $0.0$ & $80.96$ & $3.0$\\
         \hline
    \end{tabular}
    \caption{Effect of different methods for forgetting $D_f$ on VGGFace2}
    \vspace{-.1in}
    \label{tab:ablate_forget}
\end{table*}

\subsection {Results}
We present and compare our results with baseline methodologies on CIFAR-10 and VGGFace2 datasets in Table \ref{tab:vgg_results}. All results reported are the average value across 3 runs. To ensure a balanced comparison, each method is classified as either post-hoc (applied after training) or necessitating adjustments during training time. Additionally, we specify whether access to the retained dataset \(D_r\) is required for each method. Notably, our proposed method operates post-training and does not need access to \(D_r\). We report on both training and test accuracies for the retained and forgetting datasets, denoted as \(D_r\), \(D_f\), \(D_{rt}\), and \(D_{ft}\), alongside the Anamnesis Index (AIN).


We reiterate that an ideal machine unlearned is expected to forget \(D_f\) completely (i.e., low accuracies on $D_f$ and $D_{ft}$, while maintaining a high accuracy on \(D_r\) (and $D_{rt}$). In addition, an ideal unlearner must have an Anamnesis index of $1$. First, we observe that retraining the model from scratch, finetuning, and negative gradient can preserve the information on \(D_r\)  while forgetting \(D_f\) completely. However, they require access to \(D_r\), and they both have high time and memory complexity and therefore are not efficient unlearning methods. Among the methods which require access to \(D_r\), both retrain and ERM-KTP \cite{Lin_2023_CVPR} achieve the strongest results.

Our proposed method, CovarNav, excels in accuracy over ERM-KTP \cite{Lin_2023_CVPR}, while being a post-hoc approach that does not access \(D_r\). Although ERM-KTP achieves a better Anamnesis Index on VGGFace2, \(\text{AIN=7.95}\) versus our \(\text{AIN=32.02}\), CovarNav still offers competitive AINs compared to other methods that do not require \(D_r\) access. Moreover, when only accessing \(D_f\), CovarNav maintains the highest performance on both \(D_r\) and \(D_{rt}\). This robust performance on \(D_r\) is credited to our algorithm's distinctive constraint on gradient updates. Importantly, a baseline that employs the largest incorrect logit as the label proves effective in completely forgetting \(D_f\) with minimal performance impact on \(D_r\). Adding the \(\ell_2\) regularization on parameter changes provides a slightly better baseline. However, CovarNav surpasses all these methods in forgetting and retaining accuracies while maintaining a comparable AIN. 

\subsection{Ablation Studies}
In this section, we conduct various ablation studies to gain deeper insights into our proposed approach.

\subsubsection{Effect of Model Inversion}

CovarNav assumes that we do not have access to \(D_r\), and instead utilizes model inversion to obtain an approximate data set $\hat{D}_r$. A natural question arises about the effectiveness of this model inversion process. In other words, if the inverted data, $\hat{D}_r$, differs too heavily from the original \(D_r\) or does not have enough variations, we expect to be ineffective in retaining performance on \(D_r\) and \(D_{rt}\). To test this, we compare our performance to the scenario where we can access \(D_r\) for our covariance navigation versus model inversion in Table \ref{tab:ablate_inversion}. As can be seen, the model inversion can not only achieve comparable results to having full access to $D_r$ but also surprisingly provides a slightly better performance. This ablation study suggests that the inverted data represents the training data well enough to restrict the gradient space similarly. In addition, this is accomplished with a \(\hat{D}_{r}\) that is significantly smaller than \(D_r\).

\begin{figure}[b]
    \centering
    \includegraphics[width=\columnwidth]{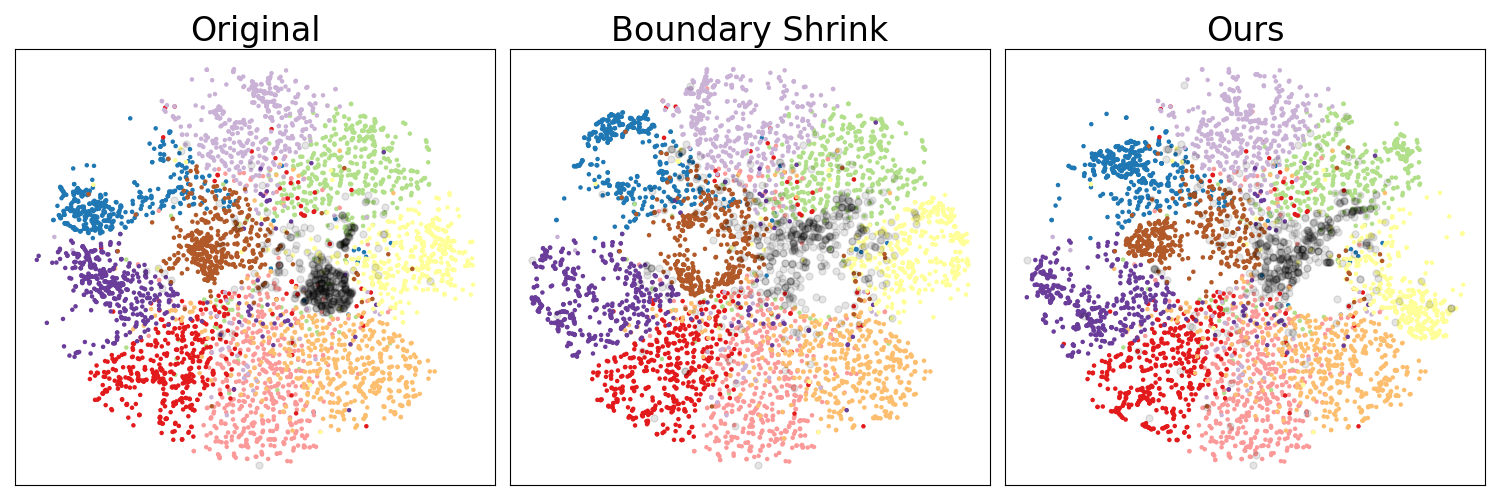}
    \caption{Joint TSNE embedding calculated based on the original model, boundary shrink, and CovarNav (Ours). The forget class is depicted with black crosses.}
    \vspace{-.2in}
    \label{fig:enter-label}
\end{figure}

\subsubsection{Effect of Forgetting Objectives on CovarNav}

In Section \ref{sec:forgetobj}, we discussed various forgetting objectives, ranging from maximizing the cross-entropy loss on \(D_f\) to employing different strategies for mislabeling \(D_f\) and minimizing the loss on this mislabeled dataset. This section investigates the impact of different forgetting objectives (as outlined in step 2 of Figure \ref{fig:overview}), in combination with our model inversion and covariance projection techniques (steps 1 and 3 in Figure \ref{fig:overview}). We examine four forgetting objectives: 1) maximizing cross-entropy, 2) mislabeling with random labels and minimizing cross-entropy, 3) boundary shrink, which involves mislabeling by identifying the closest decision boundary using FGSM, and 4) mislabeling based on the largest incorrect label. For both datasets, we report the accuracy of our method, CovarNav, on \(D_r\), \(D_{rt}\), \(D_f\), and \(D_{ft}\), utilizing these four strategies. The results are detailed in Table \ref{tab:ablate_forget}. It is observed that the strategy of using the second-largest logit consistently outperforms the other forgetting objectives.

An additional interesting point emerges when comparing Tables \ref{tab:ablate_forget} and \ref{tab:vgg_results}. Notably, the strategies of “maximizing entropy,” “random labels,” and “boundary shrink” in Table \ref{tab:vgg_results} do not incorporate covariance navigation (steps 1 and 3), while they do in Table \ref{tab:ablate_forget}. The inclusion of covariance navigation is seen to consistently enhance the performance of these methods.


\subsubsection{Embedding Visualization}
Finally, we visualized the decision boundary shifts by computing a joint TSNE embedding of the penultimate layer outputs from the original model, the model unlearned using boundary shrink, and the model unlearned using CovarNav. These results for CIFAR-10 are illustrated in Figure \ref{fig:enter-label}, where the forget class \(D_f\) is marked with black crosses. This visualization allows us to observe the effect of unlearning on the data from \(D_f\), with both boundary shrink and CovarNav methods creating a noticeable gap in the representation space previously occupied by \(D_f\). Moreover, this qualitative assessment reinforces the previously demonstrated quantitative superiority of CovarNav over boundary shrink. It is also evident that CovarNav more effectively maintains the original embedding of \(D_r\) compared to the original space than the boundary shrink method.

\section{Conclusion}
In this paper, we present a novel machine unlearning algorithm to address the need to unlearn a set of forget data, $D_f$, without having access to the retained data $D_r$. Our method consists of three steps, namely approximating the training data using model inversion, mislabeling the forget data with the largest wrong logit, and minimize the forgetting loss via projection gradient updates. We evaluate our method on CIFAR-10 and VGGFace2 datasets using accuracy and Anamnesis Index (AIN) as our metrics. Our method achieves competitive results in comparison to various state-of-the-art baselines, including boundary unlearning and ERM-KTP.

\bibliographystyle{unsrtnat}
\bibliography{main_arxiv}

\begin{thebibliography}{76}
\providecommand{\natexlab}[1]{#1}
\providecommand{\url}[1]{\texttt{#1}}
\expandafter\ifx\csname urlstyle\endcsname\relax
  \providecommand{\doi}[1]{doi: #1}\else
  \providecommand{\doi}{doi: \begingroup \urlstyle{rm}\Url}\fi

\bibitem[Fredrikson et~al.(2015)Fredrikson, Jha, and Ristenpart]{fredrikson2015model}
Matt Fredrikson, Somesh Jha, and Thomas Ristenpart.
\newblock Model inversion attacks that exploit confidence information and basic countermeasures.
\newblock In \emph{Proceedings of the 22nd ACM SIGSAC conference on computer and communications security}, pages 1322--1333, 2015.

\bibitem[Zhang et~al.(2020)Zhang, Jia, Pei, Wang, Li, and Song]{zhang2020secret}
Yuheng Zhang, Ruoxi Jia, Hengzhi Pei, Wenxiao Wang, Bo~Li, and Dawn Song.
\newblock The secret revealer: Generative model-inversion attacks against deep neural networks.
\newblock In \emph{Proceedings of the IEEE/CVF conference on computer vision and pattern recognition}, pages 253--261, 2020.

\bibitem[Shokri et~al.(2017)Shokri, Stronati, Song, and Shmatikov]{shokri2017membership}
Reza Shokri, Marco Stronati, Congzheng Song, and Vitaly Shmatikov.
\newblock Membership inference attacks against machine learning models.
\newblock In \emph{2017 IEEE symposium on security and privacy (SP)}, pages 3--18. IEEE, 2017.

\bibitem[Hu et~al.(2022)Hu, Salcic, Sun, Dobbie, Yu, and Zhang]{hu2022membership}
Hongsheng Hu, Zoran Salcic, Lichao Sun, Gillian Dobbie, Philip~S Yu, and Xuyun Zhang.
\newblock Membership inference attacks on machine learning: A survey.
\newblock \emph{ACM Computing Surveys (CSUR)}, 54\penalty0 (11s):\penalty0 1--37, 2022.

\bibitem[Regulation(2018)]{GDRP}
Protection Regulation.
\newblock General data protection regulation.
\newblock \emph{Intouch}, 25:\penalty0 1--5, 2018.

\bibitem[BUKATY(2019)]{CCPA}
PRESTON BUKATY.
\newblock \emph{The California Consumer Privacy Act (CCPA): An implementation guide}.
\newblock IT Governance Publishing, 2019.
\newblock ISBN 9781787781320.
\newblock URL \url{http://www.jstor.org/stable/j.ctvjghvnn}.

\bibitem[Jaar and Zeller(2008)]{jaar2008canadian}
Dominic Jaar and Patrick~E Zeller.
\newblock Canadian privacy law: The personal information protection and electronic documents act (pipeda).
\newblock \emph{Int'l. In-House Counsel J.}, 2:\penalty0 1135, 2008.

\bibitem[Commission et~al.(2021)]{federal2021california}
Federal~Trade Commission et~al.
\newblock California company settles ftc allegations it deceived consumers about use of facial recognition in photo storage app, 2021.

\bibitem[Cao and Yang(2015)]{cao2015towards}
Yinzhi Cao and Junfeng Yang.
\newblock Towards making systems forget with machine unlearning.
\newblock In \emph{2015 IEEE symposium on security and privacy}, pages 463--480. IEEE, 2015.

\bibitem[Golatkar et~al.(2020)Golatkar, Achille, and Soatto]{golatkar2020eternal}
Aditya Golatkar, Alessandro Achille, and Stefano Soatto.
\newblock Eternal sunshine of the spotless net: Selective forgetting in deep networks.
\newblock In \emph{Proceedings of the IEEE/CVF Conference on Computer Vision and Pattern Recognition}, pages 9304--9312, 2020.

\bibitem[Bourtoule et~al.(2021)Bourtoule, Chandrasekaran, Choquette-Choo, Jia, Travers, Zhang, Lie, and Papernot]{bourtoule2021machine}
Lucas Bourtoule, Varun Chandrasekaran, Christopher~A Choquette-Choo, Hengrui Jia, Adelin Travers, Baiwu Zhang, David Lie, and Nicolas Papernot.
\newblock Machine unlearning.
\newblock In \emph{2021 IEEE Symposium on Security and Privacy (SP)}, pages 141--159. IEEE, 2021.

\bibitem[Graves et~al.(2021)Graves, Nagisetty, and Ganesh]{graves2021amnesiac}
Laura Graves, Vineel Nagisetty, and Vijay Ganesh.
\newblock Amnesiac machine learning.
\newblock In \emph{Proceedings of the AAAI Conference on Artificial Intelligence}, volume~35, pages 11516--11524, 2021.

\bibitem[Gupta et~al.(2021)Gupta, Jung, Neel, Roth, Sharifi-Malvajerdi, and Waites]{gupta2021adaptive}
Varun Gupta, Christopher Jung, Seth Neel, Aaron Roth, Saeed Sharifi-Malvajerdi, and Chris Waites.
\newblock Adaptive machine unlearning.
\newblock \emph{Advances in Neural Information Processing Systems}, 34:\penalty0 16319--16330, 2021.

\bibitem[Sekhari et~al.(2021)Sekhari, Acharya, Kamath, and Suresh]{sekhari2021remember}
Ayush Sekhari, Jayadev Acharya, Gautam Kamath, and Ananda~Theertha Suresh.
\newblock Remember what you want to forget: Algorithms for machine unlearning.
\newblock \emph{Advances in Neural Information Processing Systems}, 34:\penalty0 18075--18086, 2021.

\bibitem[Chen et~al.(2023)Chen, Gao, Liu, Peng, and Wang]{chen2023boundary}
Min Chen, Weizhuo Gao, Gaoyang Liu, Kai Peng, and Chen Wang.
\newblock Boundary unlearning: Rapid forgetting of deep networks via shifting the decision boundary.
\newblock In \emph{Proceedings of the IEEE/CVF Conference on Computer Vision and Pattern Recognition}, pages 7766--7775, 2023.

\bibitem[Tarun et~al.(2023)Tarun, Chundawat, Mandal, and Kankanhalli]{tarun2023fast}
Ayush~K Tarun, Vikram~S Chundawat, Murari Mandal, and Mohan Kankanhalli.
\newblock Fast yet effective machine unlearning.
\newblock \emph{IEEE Transactions on Neural Networks and Learning Systems}, 2023.

\bibitem[De~Lange et~al.(2019)De~Lange, Aljundi, Masana, Parisot, Jia, Leonardis, Slabaugh, and Tuytelaars]{de2019continual}
Matthias De~Lange, Rahaf Aljundi, Marc Masana, Sarah Parisot, Xu~Jia, Ales Leonardis, Gregory Slabaugh, and Tinne Tuytelaars.
\newblock A continual learning survey: Defying forgetting in classification tasks.
\newblock \emph{arXiv preprint arXiv:1909.08383}, 2019.

\bibitem[Kudithipudi et~al.(2022)Kudithipudi, Aguilar-Simon, Babb, Bazhenov, Blackiston, Bongard, Brna, Chakravarthi~Raja, Cheney, Clune, et~al.]{kudithipudi2022biological}
Dhireesha Kudithipudi, Mario Aguilar-Simon, Jonathan Babb, Maxim Bazhenov, Douglas Blackiston, Josh Bongard, Andrew~P Brna, Suraj Chakravarthi~Raja, Nick Cheney, Jeff Clune, et~al.
\newblock Biological underpinnings for lifelong learning machines.
\newblock \emph{Nature Machine Intelligence}, 4\penalty0 (3):\penalty0 196--210, 2022.

\bibitem[Carta et~al.(2022)Carta, Cossu, Lomonaco, and Bacciu]{carta2022ex}
Antonio Carta, Andrea Cossu, Vincenzo Lomonaco, and Davide Bacciu.
\newblock Ex-model: Continual learning from a stream of trained models.
\newblock In \emph{Proceedings of the IEEE/CVF Conference on Computer Vision and Pattern Recognition}, pages 3790--3799, 2022.

\bibitem[Yan et~al.(2022)Yan, Hong, Xu, Han, Tuytelaars, Li, and He]{yan2022generative}
Shipeng Yan, Lanqing Hong, Hang Xu, Jianhua Han, Tinne Tuytelaars, Zhenguo Li, and Xuming He.
\newblock Generative negative text replay for continual vision-language pretraining.
\newblock In \emph{European Conference on Computer Vision}, pages 22--38. Springer, 2022.

\bibitem[Madaan et~al.(2023)Madaan, Yin, Byeon, Kautz, and Molchanov]{madaan2023heterogeneous}
Divyam Madaan, Hongxu Yin, Wonmin Byeon, Jan Kautz, and Pavlo Molchanov.
\newblock Heterogeneous continual learning.
\newblock In \emph{Proceedings of the IEEE/CVF Conference on Computer Vision and Pattern Recognition}, pages 15985--15995, 2023.

\bibitem[Saha et~al.(2020)Saha, Garg, and Roy]{saha2020gradient}
Gobinda Saha, Isha Garg, and Kaushik Roy.
\newblock Gradient projection memory for continual learning.
\newblock In \emph{International Conference on Learning Representations}, 2020.

\bibitem[Wang et~al.(2021{\natexlab{a}})Wang, Li, Sun, and Xu]{wang2021covarnull}
Shipeng Wang, Xiaorong Li, Jian Sun, and Zongben Xu.
\newblock Training networks in null space of feature covariance for continual learning.
\newblock In \emph{Proceedings of the IEEE/CVF conference on Computer Vision and Pattern Recognition}, pages 184--193, 2021{\natexlab{a}}.

\bibitem[Abbasi et~al.(2022)Abbasi, Nooralinejad, Braverman, Pirsiavash, and Kolouri]{abbasi2022sparsity}
Ali Abbasi, Parsa Nooralinejad, Vladimir Braverman, Hamed Pirsiavash, and Soheil Kolouri.
\newblock Sparsity and heterogeneous dropout for continual learning in the null space of neural activations.
\newblock In \emph{Conference on Lifelong Learning Agents}, pages 617--628. PMLR, 2022.

\bibitem[Yin et~al.(2020)Yin, Molchanov, Alvarez, Li, Mallya, Hoiem, Jha, and Kautz]{yin2020dreaming}
Hongxu Yin, Pavlo Molchanov, Jose~M Alvarez, Zhizhong Li, Arun Mallya, Derek Hoiem, Niraj~K Jha, and Jan Kautz.
\newblock Dreaming to distill: Data-free knowledge transfer via deepinversion.
\newblock In \emph{Proceedings of the IEEE/CVF Conference on Computer Vision and Pattern Recognition}, pages 8715--8724, 2020.

\bibitem[Wang et~al.(2021{\natexlab{b}})Wang, Fu, Li, Khisti, Zemel, and Makhzani]{wang2021variational}
Kuan-Chieh Wang, Yan Fu, Ke~Li, Ashish Khisti, Richard Zemel, and Alireza Makhzani.
\newblock Variational model inversion attacks.
\newblock \emph{Advances in Neural Information Processing Systems}, 34:\penalty0 9706--9719, 2021{\natexlab{b}}.

\bibitem[Xu et~al.(2023)Xu, Zhu, Zhang, Zhou, and Yu]{xu2023machine}
Heng Xu, Tianqing Zhu, Lefeng Zhang, Wanlei Zhou, and Philip~S Yu.
\newblock Machine unlearning: A survey.
\newblock \emph{ACM Computing Surveys}, 56\penalty0 (1):\penalty0 1--36, 2023.

\bibitem[Zhang et~al.(2023{\natexlab{a}})Zhang, Nakamura, Isohara, and Sakurai]{zhang2023review}
Haibo Zhang, Toru Nakamura, Takamasa Isohara, and Kouichi Sakurai.
\newblock A review on machine unlearning.
\newblock \emph{SN Computer Science}, 4\penalty0 (4):\penalty0 337, 2023{\natexlab{a}}.

\bibitem[Nguyen et~al.(2022)Nguyen, Huynh, Nguyen, Liew, Yin, and Nguyen]{nguyen2022survey}
Thanh~Tam Nguyen, Thanh~Trung Huynh, Phi~Le Nguyen, Alan Wee-Chung Liew, Hongzhi Yin, and Quoc Viet~Hung Nguyen.
\newblock A survey of machine unlearning.
\newblock \emph{arXiv preprint arXiv:2209.02299}, 2022.

\bibitem[Cauwenberghs and Poggio(2000)]{cauwenberghs2000incremental}
Gert Cauwenberghs and Tomaso Poggio.
\newblock Incremental and decremental support vector machine learning.
\newblock \emph{Advances in neural information processing systems}, 13, 2000.

\bibitem[Romero et~al.(2007)Romero, Barrio, and Belanche]{romero2007incremental}
Enrique Romero, Ignacio Barrio, and Llu{\'\i}s Belanche.
\newblock Incremental and decremental learning for linear support vector machines.
\newblock In \emph{International Conference on Artificial Neural Networks}, pages 209--218. Springer, 2007.

\bibitem[Tsai et~al.(2014)Tsai, Lin, and Lin]{tsai2014incremental}
Cheng-Hao Tsai, Chieh-Yen Lin, and Chih-Jen Lin.
\newblock Incremental and decremental training for linear classification.
\newblock In \emph{Proceedings of the 20th ACM SIGKDD international conference on Knowledge discovery and data mining}, pages 343--352, 2014.

\bibitem[Cao et~al.(2018)Cao, Yu, Aday, Stahl, Merwine, and Yang]{cao2018efficient}
Yinzhi Cao, Alexander~Fangxiao Yu, Andrew Aday, Eric Stahl, Jon Merwine, and Junfeng Yang.
\newblock Efficient repair of polluted machine learning systems via causal unlearning.
\newblock In \emph{Proceedings of the 2018 on Asia conference on computer and communications security}, pages 735--747, 2018.

\bibitem[Brophy and Lowd(2021)]{brophy2021machine}
Jonathan Brophy and Daniel Lowd.
\newblock Machine unlearning for random forests.
\newblock In \emph{International Conference on Machine Learning}, pages 1092--1104. PMLR, 2021.

\bibitem[Liu et~al.(2021)Liu, Ma, Yang, Wang, and Liu]{liu2021federaser}
Gaoyang Liu, Xiaoqiang Ma, Yang Yang, Chen Wang, and Jiangchuan Liu.
\newblock Federaser: Enabling efficient client-level data removal from federated learning models.
\newblock In \emph{2021 IEEE/ACM 29th International Symposium on Quality of Service (IWQOS)}, pages 1--10. IEEE, 2021.

\bibitem[Jose and Simeone(2021)]{jose2021unified}
Sharu~Theresa Jose and Osvaldo Simeone.
\newblock A unified pac-bayesian framework for machine unlearning via information risk minimization.
\newblock In \emph{2021 IEEE 31st International Workshop on Machine Learning for Signal Processing (MLSP)}, pages 1--6. IEEE, 2021.

\bibitem[Kashef(2021)]{kashef2021boosted}
Rasha Kashef.
\newblock A boosted svm classifier trained by incremental learning and decremental unlearning approach.
\newblock \emph{Expert Systems with Applications}, 167:\penalty0 114154, 2021.

\bibitem[Ullah et~al.(2021)Ullah, Mai, Rao, Rossi, and Arora]{ullah2021machine}
Enayat Ullah, Tung Mai, Anup Rao, Ryan~A Rossi, and Raman Arora.
\newblock Machine unlearning via algorithmic stability.
\newblock In \emph{Conference on Learning Theory}, pages 4126--4142. PMLR, 2021.

\bibitem[Neel et~al.(2021)Neel, Roth, and Sharifi-Malvajerdi]{neel2021descent}
Seth Neel, Aaron Roth, and Saeed Sharifi-Malvajerdi.
\newblock Descent-to-delete: Gradient-based methods for machine unlearning.
\newblock In \emph{Algorithmic Learning Theory}, pages 931--962. PMLR, 2021.

\bibitem[Thudi et~al.(2022)Thudi, Deza, Chandrasekaran, and Papernot]{thudi2022unrolling}
Anvith Thudi, Gabriel Deza, Varun Chandrasekaran, and Nicolas Papernot.
\newblock Unrolling sgd: Understanding factors influencing machine unlearning.
\newblock In \emph{2022 IEEE 7th European Symposium on Security and Privacy (EuroS\&P)}, pages 303--319. IEEE, 2022.

\bibitem[De~Lange et~al.(2021)De~Lange, Aljundi, Masana, Parisot, Jia, Leonardis, Slabaugh, and Tuytelaars]{de2021continual}
Matthias De~Lange, Rahaf Aljundi, Marc Masana, Sarah Parisot, Xu~Jia, Ale{\v{s}} Leonardis, Gregory Slabaugh, and Tinne Tuytelaars.
\newblock A continual learning survey: Defying forgetting in classification tasks.
\newblock \emph{IEEE transactions on pattern analysis and machine intelligence}, 44\penalty0 (7):\penalty0 3366--3385, 2021.

\bibitem[Shin et~al.(2017)Shin, Lee, Kim, and Kim]{shin2017continual}
Hanul Shin, Jung~Kwon Lee, Jaehong Kim, and Jiwon Kim.
\newblock Continual learning with deep generative replay.
\newblock In \emph{Proceedings of the 31st International Conference on Neural Information Processing Systems}, pages 2994--3003, 2017.

\bibitem[Farquhar and Gal(2018)]{farquhar2018towards}
Sebastian Farquhar and Yarin Gal.
\newblock Towards robust evaluations of continual learning.
\newblock \emph{arXiv preprint arXiv:1805.09733}, 2018.

\bibitem[Rolnick et~al.(2019)Rolnick, Ahuja, Schwarz, Lillicrap, and Wayne]{rolnick2019experience}
David Rolnick, Arun Ahuja, Jonathan Schwarz, Timothy Lillicrap, and Gregory Wayne.
\newblock Experience replay for continual learning.
\newblock \emph{Advances in Neural Information Processing Systems}, 32, 2019.

\bibitem[Rostami et~al.(2020)Rostami, Kolouri, Pilly, and McClelland]{rostami2020generative}
Mohammad Rostami, Soheil Kolouri, Praveen Pilly, and James McClelland.
\newblock Generative continual concept learning.
\newblock In \emph{Proceedings of the AAAI Conference on Artificial Intelligence}, volume~34, pages 5545--5552, 2020.

\bibitem[Farajtabar et~al.(2020)Farajtabar, Azizan, Mott, and Li]{farajtabar2020orthogonal}
Mehrdad Farajtabar, Navid Azizan, Alex Mott, and Ang Li.
\newblock Orthogonal gradient descent for continual learning.
\newblock In \emph{International Conference on Artificial Intelligence and Statistics}, pages 3762--3773. PMLR, 2020.

\bibitem[Kirkpatrick et~al.(2017)Kirkpatrick, Pascanu, Rabinowitz, Veness, Desjardins, Rusu, Milan, Quan, Ramalho, Grabska-Barwinska, et~al.]{kirkpatrick2017overcoming}
James Kirkpatrick, Razvan Pascanu, Neil Rabinowitz, Joel Veness, Guillaume Desjardins, Andrei~A Rusu, Kieran Milan, John Quan, Tiago Ramalho, Agnieszka Grabska-Barwinska, et~al.
\newblock Overcoming catastrophic forgetting in neural networks.
\newblock \emph{Proceedings of the national academy of sciences}, 114\penalty0 (13):\penalty0 3521--3526, 2017.

\bibitem[Zenke et~al.(2017)Zenke, Poole, and Ganguli]{zenke2017continual}
Friedemann Zenke, Ben Poole, and Surya Ganguli.
\newblock Continual learning through synaptic intelligence.
\newblock In \emph{International Conference on Machine Learning}, pages 3987--3995. PMLR, 2017.

\bibitem[Aljundi et~al.(2018)Aljundi, Babiloni, Elhoseiny, Rohrbach, and Tuytelaars]{aljundi2017memory}
Rahaf Aljundi, Francesca Babiloni, Mohamed Elhoseiny, Marcus Rohrbach, and Tinne Tuytelaars.
\newblock Memory aware synapses: Learning what (not) to forget.
\newblock In \emph{ECCV}, 2018.

\bibitem[Kolouri et~al.(2020)Kolouri, Ketz, Soltoggio, and Pilly]{kolouri2020sliced}
Soheil Kolouri, Nicholas~A Ketz, Andrea Soltoggio, and Praveen~K Pilly.
\newblock Sliced cramer synaptic consolidation for preserving deeply learned representations.
\newblock In \emph{International Conference on Learning Representations}, 2020.

\bibitem[li2(2021)]{li2021lifelong}
\emph{Lifelong Learning with Sketched Structural Regularization}, volume 157 of \emph{Proceedings of Machine Learning Research}, 2021. {PMLR}.

\bibitem[von Oswald et~al.(2019)von Oswald, Henning, Sacramento, and Grewe]{von2019continual}
Johannes von Oswald, Christian Henning, Jo{\~a}o Sacramento, and Benjamin~F Grewe.
\newblock Continual learning with hypernetworks.
\newblock \emph{arXiv preprint arXiv:1906.00695}, 2019.

\bibitem[Schwarz et~al.(2018)Schwarz, Czarnecki, Luketina, Grabska-Barwinska, Teh, Pascanu, and Hadsell]{schwarz2018progress}
Jonathan Schwarz, Wojciech Czarnecki, Jelena Luketina, Agnieszka Grabska-Barwinska, Yee~Whye Teh, Razvan Pascanu, and Raia Hadsell.
\newblock Progress \& compress: A scalable framework for continual learning.
\newblock In \emph{International Conference on Machine Learning}, pages 4528--4537. PMLR, 2018.

\bibitem[Mallya and Lazebnik(2018)]{mallya2018packnet}
Arun Mallya and Svetlana Lazebnik.
\newblock Packnet: Adding multiple tasks to a single network by iterative pruning.
\newblock In \emph{Proceedings of the IEEE conference on Computer Vision and Pattern Recognition}, pages 7765--7773, 2018.

\bibitem[Mallya et~al.(2018)Mallya, Davis, and Lazebnik]{mallya2018piggyback}
Arun Mallya, Dillon Davis, and Svetlana Lazebnik.
\newblock Piggyback: Adapting a single network to multiple tasks by learning to mask weights.
\newblock In \emph{Proceedings of the European Conference on Computer Vision (ECCV)}, pages 67--82, 2018.

\bibitem[Wortsman et~al.(2020)Wortsman, Ramanujan, Liu, Kembhavi, Rastegari, Yosinski, and Farhadi]{wortsman2020supermasks}
Mitchell Wortsman, Vivek Ramanujan, Rosanne Liu, Aniruddha Kembhavi, Mohammad Rastegari, Jason Yosinski, and Ali Farhadi.
\newblock Supermasks in superposition.
\newblock \emph{Advances in Neural Information Processing Systems}, 33:\penalty0 15173--15184, 2020.

\bibitem[Ben-Iwhiwhu et~al.(2023)Ben-Iwhiwhu, Nath, Pilly, Kolouri, and Soltoggio]{ben-iwhiwhu2023lifelong}
Eseoghene Ben-Iwhiwhu, Saptarshi Nath, Praveen~Kumar Pilly, Soheil Kolouri, and Andrea Soltoggio.
\newblock Lifelong reinforcement learning with modulating masks.
\newblock \emph{Transactions on Machine Learning Research}, 2023.
\newblock ISSN 2835-8856.
\newblock URL \url{https://openreview.net/forum?id=V7tahqGrOq}.

\bibitem[Deng et~al.(2021)Deng, Chen, HAO, Wang, and Heng]{deng2021flattening}
Danruo Deng, Guangyong Chen, Jianye HAO, Qiong Wang, and Pheng-Ann Heng.
\newblock Flattening sharpness for dynamic gradient projection memory benefits continual learning.
\newblock In A.~Beygelzimer, Y.~Dauphin, P.~Liang, and J.~Wortman Vaughan, editors, \emph{Advances in Neural Information Processing Systems}, 2021.
\newblock URL \url{https://openreview.net/forum?id=q1eCa1kMfDd}.

\bibitem[Wang et~al.(2021{\natexlab{c}})Wang, Li, Sun, and Xu]{wang2021training}
Shipeng Wang, Xiaorong Li, Jian Sun, and Zongben Xu.
\newblock Training networks in null space of feature covariance for continual learning.
\newblock In \emph{Proceedings of the IEEE/CVF Conference on Computer Vision and Pattern Recognition}, pages 184--193, 2021{\natexlab{c}}.

\bibitem[Lin et~al.(2022)Lin, Yang, Fan, and Zhang]{lin2022trgp}
Sen Lin, Li~Yang, Deliang Fan, and Junshan Zhang.
\newblock Trgp: Trust region gradient projection for continual learning.
\newblock In \emph{International Conference on Learning Representations}, 2022.

\bibitem[Zhao et~al.(2023)Zhao, Zhang, Tan, Liu, Qu, Xie, and Ma]{zhao2023CVPR}
Zhen Zhao, Zhizhong Zhang, Xin Tan, Jun Liu, Yanyun Qu, Yuan Xie, and Lizhuang Ma.
\newblock Rethinking gradient projection continual learning: Stability / plasticity feature space decoupling.
\newblock In \emph{Proceedings of the IEEE/CVF Conference on Computer Vision and Pattern Recognition (CVPR)}, pages 3718--3727, June 2023.

\bibitem[Shibata et~al.(2021)Shibata, Irie, Ikami, and Mitsuzumi]{shibata2021learning}
Takashi Shibata, Go~Irie, Daiki Ikami, and Yu~Mitsuzumi.
\newblock Learning with selective forgetting.
\newblock In \emph{IJCAI}, volume~3, page~4, 2021.

\bibitem[Heng and Soh(2023)]{heng2023selective}
Alvin Heng and Harold Soh.
\newblock Selective amnesia: A continual learning approach to forgetting in deep generative models.
\newblock \emph{Advances in Neural Information Processing Systems}, 2023.

\bibitem[Zhang et~al.(2023{\natexlab{b}})Zhang, Lu, Zhang, Wang, and Li]{zhang2023machine}
Yongjing Zhang, Zhaobo Lu, Feng Zhang, Hao Wang, and Shaojing Li.
\newblock Machine unlearning by reversing the continual learning.
\newblock \emph{Applied Sciences}, 13\penalty0 (16):\penalty0 9341, 2023{\natexlab{b}}.

\bibitem[Veale et~al.(2018)Veale, Binns, and Edwards]{veale2018algorithms}
Michael Veale, Reuben Binns, and Lilian Edwards.
\newblock Algorithms that remember: model inversion attacks and data protection law.
\newblock \emph{Philosophical Transactions of the Royal Society A: Mathematical, Physical and Engineering Sciences}, 376\penalty0 (2133):\penalty0 20180083, 2018.

\bibitem[Mahendran and Vedaldi(2015)]{mahendran2015understanding}
Aravindh Mahendran and Andrea Vedaldi.
\newblock Understanding deep image representations by inverting them.
\newblock In \emph{Proceedings of the IEEE conference on computer vision and pattern recognition}, pages 5188--5196, 2015.

\bibitem[Mordvintsev et~al.(2015)Mordvintsev, Olah, and Tyka]{mordvintsev2015deepdream}
Alexander Mordvintsev, Christopher Olah, and Mike Tyka.
\newblock Deepdream-a code example for visualizing neural networks.
\newblock \emph{Google Research}, 2\penalty0 (5), 2015.

\bibitem[Chundawat et~al.(2023)Chundawat, Tarun, Mandal, and Kankanhalli]{chundawat2023zero}
Vikram~S Chundawat, Ayush~K Tarun, Murari Mandal, and Mohan Kankanhalli.
\newblock Zero-shot machine unlearning.
\newblock \emph{IEEE Transactions on Information Forensics and Security}, 2023.

\bibitem[Ioffe and Szegedy(2015)]{ioffe2015batch}
Sergey Ioffe and Christian Szegedy.
\newblock Batch normalization: Accelerating deep network training by reducing internal covariate shift.
\newblock In \emph{International conference on machine learning}, pages 448--456. pmlr, 2015.

\bibitem[Goodfellow et~al.(2014)Goodfellow, Shlens, and Szegedy]{goodfellow2014explaining}
Ian~J Goodfellow, Jonathon Shlens, and Christian Szegedy.
\newblock Explaining and harnessing adversarial examples.
\newblock \emph{arXiv preprint arXiv:1412.6572}, 2014.

\bibitem[Lee and Woo(2023)]{lee2023undo}
Sangyong Lee and Simon~S Woo.
\newblock Undo: Effective and accurate unlearning method for deep neural networks.
\newblock In \emph{Proceedings of the 32nd ACM International Conference on Information and Knowledge Management}, pages 4043--4047, 2023.

\bibitem[Krizhevsky et~al.(2009)Krizhevsky, Hinton, et~al.]{krizhevsky2009learning}
Alex Krizhevsky, Geoffrey Hinton, et~al.
\newblock Learning multiple layers of features from tiny images.
\newblock 2009.

\bibitem[Cao et~al.(2017)Cao, Shen, Xie, Parkhi, and Zisserman]{DBLP:journals/corr/abs-1710-08092}
Qiong Cao, Li~Shen, Weidi Xie, Omkar~M. Parkhi, and Andrew Zisserman.
\newblock Vggface2: {A} dataset for recognising faces across pose and age.
\newblock \emph{CoRR}, abs/1710.08092, 2017.
\newblock URL \url{http://arxiv.org/abs/1710.08092}.

\bibitem[Lin et~al.(2023)Lin, Zhang, Chen, Chen, and Susilo]{Lin_2023_CVPR}
Shen Lin, Xiaoyu Zhang, Chenyang Chen, Xiaofeng Chen, and Willy Susilo.
\newblock Erm-ktp: Knowledge-level machine unlearning via knowledge transfer.
\newblock In \emph{Proceedings of the IEEE/CVF Conference on Computer Vision and Pattern Recognition (CVPR)}, pages 20147--20155, June 2023.

\bibitem[He et~al.(2015)He, Zhang, Ren, and Sun]{He2015}
Kaiming He, Xiangyu Zhang, Shaoqing Ren, and Jian Sun.
\newblock Deep residual learning for image recognition.
\newblock \emph{arXiv preprint arXiv:1512.03385}, 2015.

\bibitem[Kingma and Ba(2015)]{KingBa15}
Diederik Kingma and Jimmy Ba.
\newblock Adam: A method for stochastic optimization.
\newblock In \emph{International Conference on Learning Representations (ICLR)}, San Diega, CA, USA, 2015.

\end{thebibliography}

\end{document}